\documentclass{article}

\usepackage{microtype}
\usepackage{graphicx}
\usepackage{booktabs} % for professional tables

\usepackage{caption}
\usepackage{multirow}
\usepackage{multicol}
\usepackage{setspace}
\usepackage[hang, tight]{subfigure}
\usepackage{url}
\usepackage{verbatim}
\usepackage{wrapfig}
\usepackage[usenames,dvipsnames]
{xcolor}

\usepackage{algorithmic}
\usepackage{makecell}
\usepackage{algorithm}
\usepackage{amsmath}
\usepackage{amssymb}

\usepackage{array}
\usepackage{booktabs}
\usepackage[bookmarks=false]{hyperref}
\usepackage{caption}
\usepackage{color}
\usepackage{comment}
\usepackage[inline]{enumitem}
\usepackage{epsfig}
\usepackage{etoolbox}
\usepackage{fancybox}
\usepackage{float}
\usepackage{fixltx2e}
\usepackage{graphicx}

\usepackage{xcolor}
\usepackage{soul}
\definecolor{Seashell}{RGB}{0, 0, 0} %背景色浅一点的
\definecolor{Firebrick4}{RGB}{255, 255, 255}%文字颜色红一点的

\newcommand{\code}[1]{
  \begingroup
  \sethlcolor{Seashell}%背景色
  \textcolor{Firebrick4}{\hl{#1}}%textcolor里面对应文字颜色
  \endgroup
}

\usepackage{listings}
\usepackage{color}

\definecolor{dkgreen}{rgb}{0,0.6,0}
\definecolor{gray}{rgb}{0.5,0.5,0.5}
\definecolor{mauve}{rgb}{0.58,0,0.82}

\usepackage[accepted]{mlsys2022}

\mlsystitlerunning{QuadraLib: A Performant Quadratic Neural Network Library for Architecture Optimization and Design Exploration}

\graphicspath{{}}

\begin{document}

\twocolumn[

\mlsystitle{QuadraLib: A Performant Quadratic Neural Network Library \\for Architecture Optimization and Design Exploration}

\mlsyssetsymbol{equal}{*}

\begin{mlsysauthorlist}
\mlsysauthor{Zirui Xu}{to}
\mlsysauthor{Fuxun Yu}{to}
\mlsysauthor{Jinjun Xiong}{goo}
\mlsysauthor{Xiang Chen}{to}

\end{mlsysauthorlist}

\mlsysaffiliation{to}{George Mason University, VA, USA}
\mlsysaffiliation{goo}{University at Buffalo, NY, USA}
% \mlsysaffiliation{ed}{School of Computation, University of Edenborrow, Edenborrow, United Kingdom}

\mlsyscorrespondingauthor{Xiang Chen}{xchen26@gmu.edu}
\mlsyscorrespondingauthor{Jinjun Xiong}{jinjun@buffalo.edu}

\mlsyskeywords{Machine Learning, MLSys}

\vskip 0.3in

% \begin{abstract}
% This document provides a basic paper template and submission guidelines.
% Abstracts must be a single paragraph, ideally between 4--6 sentences long.
% Gross violations will trigger corrections at the camera-ready phase.
% \end{abstract}
\begin{abstract}

% Deep Neural Networks (DNNs) have been widely used in various learning scenarios with huge success. Behind it, 

% The emergence of Deep Learning (DL) frameworks significantly si

% DNN 在各种application大规模使用并且取得了显著成功，我们发现这其中除了算法的革新，更需要归功于DNN framework。他们enable user 可以方便地构建model，efficiently去train/inference model, 并且提供了analysis tool去实现更多的numerical analysis 和 exploration。QDNN 被证明比DNN有更强的nonlinearity和approximation capability同时更efficient。但是现在没有一个QDNN linrary来support various QDNN design 并且进行分析。
% 为此我们提出了quadralib，通过对现有design的分析我们从theoretical的角度提出了一个更好的neuron design。然后我们的library能进行所有existing work的numerical analysis, 还提供了trianing 优化和几个tool实现model building和analysis。

The significant success of Deep Neural Networks (DNNs) is highly promoted by the multiple sophisticated DNN libraries. On the contrary, although some work have proved that Quadratic Deep Neuron Networks (QDNNs) show better non-linearity and learning capability than the first-order DNNs, their neuron design suffers certain drawbacks from theoretical performance to practical deployment. In this paper, we first proposed a new QDNN neuron architecture design, and further developed \textit{QuadraLib}, a QDNN library to provide architecture optimization and design exploration for QDNNs. Extensive experiments show that our design has good performance regarding prediction accuracy and computation consumption on multiple learning tasks.

\end{abstract}

]

\printAffiliationsAndNotice{}  % leave blank if no need to mention equal contribution
% \printAffiliationsAndNotice{\mlsysEqualContribution} % otherwise use the standard text.

% \input{_txt/0_abstract}
\section{Introduction}

% 用词统一：1）linear neuron， quadratic neuron，2）first-order term/second-order term  3) first-order DNNs, QDNNs  4) first-order polynomial/ second-order polynomial

The availability of easy-to-use Deep Learning (DL) frameworks/libraries, ranging from the early days' Theano~\cite{bergstra2010theano}, Caffee~\cite{jia2014caffe}, Torch~\cite{paszke2019pytorch} to the 
recent Tensorflow~\cite{abadi2016tensorflow}, PyTorch~\cite{paszke2019pytorch}, MxNet~\cite{chen2015mxnet}, etc. have undoubtedly contributed to the popularity of DL and the proliferation of many successful deep neural networks (DNNs) for a variety of applications, such as \textit{VGGNet}~\cite{simonyan2014very} and \textit{ResNet}~\cite{he2016deep} for image classification,
\textit{Faster RCNN}~\cite{ren2015faster} and \textit{SSD}~\cite{liu2016ssd} for object detection, and \textit{BERT}~\cite{devlin2019bert} and \textit{GPT-3}~\cite{brown2020language} for natural language modeling. The success and wide adoption of these DNN models in turn raise many new requirements for these DL frameworks, thus pushing the continued development of new functionalities inside
these DL frameworks. This has created a positive feedback cycle for the unparalleled
prosperity of the entire DL ecosystem, and has become a new source of inspiration for innovation for the entire machine learning community.

\begin{figure}[!t]
	\centering
	\captionsetup{justification=centering}
	\vspace{0mm}
	\includegraphics[width=3.3in]{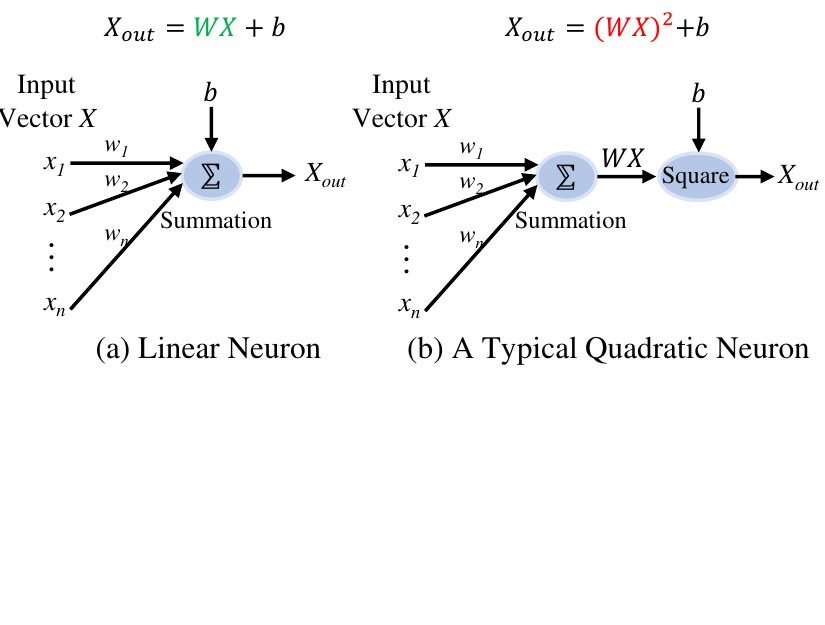}
	\vspace{-34mm}
	\caption{From Linear Neuron to Quadratic Neuron}
	\vspace{-9mm}
	\label{CNN_to_QCNN}
\end{figure}

One recent notable progress in improving DNN's learning capacity is the proposal of Quadratic Deep Neural Networks (QDNNs). In contrast
to the existing DNNs where each neuron is represented by a linear combination of its inputs $X$ and weight parameters $W$ (i.e.,
a first-order polynomial form) as shown in Fig.~\ref{CNN_to_QCNN} (a), the QDNN's neurons are represented by a second-order polynomial
of inputs $X$ and weight parameters $W$ as shown in Fig.~\ref{CNN_to_QCNN} (b), thus improving DNN models' learning capacity 
~\cite{milenkovic1996annealing,redlapalli2003development,zoumpourlis2017non,ganesh2017pattern,fan2018new,
jiang2019nonlinear,brooks2019second,mantinicqnn,chrysos2020p,bu2021quadratic}.         Compared to the linear neurons, the benefits of QDNNs stem from the
unique characteristics of the second-order polynomial form: (1) stronger non-linearity, hence
improved capability for feature extraction~\cite{jiang2019nonlinear}, and (2) higher model efficiency as QDNN can approximate polynomial decision boundaries using smaller network depth/width~\cite{chrysos2020p, bu2021quadratic}.
Moreover, with such non-linearity, replacing
ReLU with quadratic layer will significantly reduce the computation cost in many Privacy-Preserving Machine Learning (PPML) protocol designs~\cite{mishra2020delphi, gilad2016cryptonets, chou2018faster, garimella2021sisyphus}.
% popular-used  two-party (2PC) computation-based cryption protocols. 
% and (3) \ZX{We may need to remove the third one?} improved model capacity, as QDNN can be expanded to encompass arbitrary higher order polynomials without relying on nonlinear activation functions (which is proven useful for applications with high polynomial mapping functions, such as image generation~\cite{chrysos2020p}). 
In spite of these great potentials, the 
number of follow-up works in QDNNs is, however, substantially smaller than the traditional first-order DNNs.

% issues: library以前的work只能手动编写，只有一些自定义的简单结构，  后来user-friendly，方便model building，产生了公认的dnn结构，再后来有了model zoom, analysis tools，
% 以前没有user API，智能手动编写
% 后来有了各种pre-define的DNN model，
% 后来有了trianing optimization还有分析工具

\begin{figure}[!t]
	\centering
	\captionsetup{justification=centering}
	\vspace{-2mm}
	\includegraphics[width=3.3in]{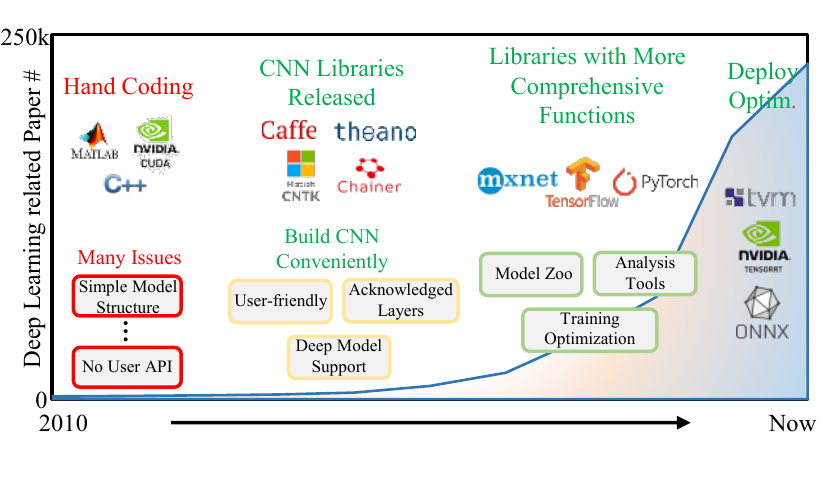}
	\vspace{-14mm}
	\caption{DNN Library Timeline and Paper Number}
	\vspace{-6mm}
	\label{paper_number_2}
\end{figure}

When we look closely at the evolution path for the first-order DNNs' development as shown in Fig.~\ref{paper_number_2}, 
it has seen a similar issue. For example, although Convolutional Neural Networks (CNNs) were proposed as early as 1990s, there were few follow-up works until around 2012. The major roadblock was the productivity issue as researchers during that period had to build models by writing customized software with low-level primitives (such as C/C$++$ and Nvidia CUDA). 
Starting around 2012, DNN development started to ramp up with the introduction of DNN libraries and frameworks as pioneered by 
Theano~\cite{bergstra2010theano}, Torch~\cite{torch}, 
and Chainer~\cite{tokui2015chainer} etc.
These works provided a user-friendly APIs to include a set of
pre-defined DNN layers and models for network construction, and the number of DNN-based research has since increased significantly as represented by the volume of published research papers around DNNs.  
The diversity of DNN/CNN models raised the need for even more productive tools and libraries, which incentivised companies to open-source frameworks like Tensorflow~\cite{abadi2016tensorflow}, Pytorch~\cite{paszke2019pytorch}, etc. These libraries further provided optimized training/inference computation to ensure fast and optimal convergence of new DNN models.
Moreover, some model visualization tools were developed to enable in-depth analyses of DNN models, such as TensorBoard~\cite{abadi2016tensorflow}.  
In summary, as demonstrated in Fig.~\ref{paper_number_2}, a set of well-defined
libraries and tools really enabled the research community to fully understand the power of first-order DNN by facilitating their fast model exploration, thus speeding up their innovations in designing
more powerful DNN models for the past decade.

Inspired by this historical success for the first-order DNN development, we believe
a similar effort is needed for the second-order DNNs, i.e., QDNNs, so that we can enable more researchers to better explore various QDNN models and understand QDNNs' strengths and weakness. The contributions of our work are as follows:

\begin{itemize}
% \item We first survey the exist QDNN work in the literature and categorize the various QDNN design into four types based on their key neuron computation format, which helps us to extract the common computation patterns in order to support various QDNN models.
\vspace{-3mm}
\item We first categorized the existing QDNN design into four types based on their neuron architecture and then comprehensively analyzed the drawbacks of each design type from multiple perspectives. 
% \item We first survey the exist QDNN design in the literature and categorize  them into four types based on their key neuron computation format. By investigating their various computation patterns, %from multi-aspects, 
% we propose a unified neuron design which is high performant for QDNN's practical deployment.

\vspace{-2mm}
\item Based on the drawbacks analysis, we proposed a new quadratic neuron architecture. The effectiveness and efficiency of the proposed neuron is further analyzed from the theoretical aspect.

\vspace{-2mm}
\item We developed a second-order DNN library based on PyTorch, called \textit{QuadraLib},
% that not only supports the common computation patterns for the existing QDNNs but also can conveniently construct QDNN structure with optimal performance.  
% that not only supports flexible model construction for various QDNNs but also optimizes memory overhead so that QDNN training can be of high performance as well. 
that not only supports flexible model construction for various QDNNs but also optimizes memory overhead for QDNN training. 

% \vspace{-2mm}
% \item The proposed \textit{QuadraLib} can also optimize memory overhead for carrying out QDNN computations so that QDNN training can have high performance as well.
% \item We develop a second-order DNN library based on Pytorch, called \textit{QuadraLib},
% that not only supports the common computation patterns for existing QDNNs and our new DNN design, but also optimizes memory overhead for carrying out all QDNN computations so that QDNN training can be of high performance as well.

% \vspace{-2mm}
% \item We further proposed a QDNN model converter in \textit{QuadraLib}. Through such converter, users can obtain an optimal QDNN model based on a given first-order DNN structure with minimum effort.

% We further develop two productive tools at the top-API level of \textit{QuadraLib}, which can help users quickly generate their personalized QDNN models and conduct in-depth network analysis. %according to activation and weight/gradient. 

\vspace{-2mm}
\item We conducted a systematic study of our proposed QDNN model on various learning tasks, including image classification, image generation and object detection. The extensive experimental results validate its superiority compared to the first-order DNNs and the existing QDNN models.
% All of experimentation is made possible  because of \textit{QuadraLib}. 
\end{itemize}

\vspace{-4mm}
We plan to open source our \textit{QuadraLib} to the community with a simple-to-use installation instruction. 
We hope this will stimulate the community's interests in conducting more active research around QDNNs so that we can jointly expand the arsenal of the existing DNN models to include higher-order neuron networks and also provide potential benefits to PPML society.

\begin{table*}[t]
\small
\renewcommand\arraystretch{1.5}
\centering
\vspace{0mm}
\caption{The Overview of Current QDNN Works}
\vspace{2mm}
\setlength{\tabcolsep}{0.6mm}{
\begin{tabular}{|c|c|c||c|c||c||c|c|}
\hline
\begin{tabular}[c]{@{}c@{}}Type \vspace{-1mm} \\  \end{tabular} & \begin{tabular}[c]{@{}c@{}}Neuron  \vspace{-1mm}\\ Format \end{tabular}                          & \begin{tabular}[c]{@{}c@{}}Work \vspace{-1mm} \\ Reference \end{tabular} & \multicolumn{2}{c|}{\begin{tabular}[c]{@{}c@{}}Computation  \vspace{-1mm}\\ Complexity\end{tabular}}   &\begin{tabular}[c]{@{}c@{}}Model  \vspace{-1mm} \\ Structure$^{\dagger}$ \end{tabular}      & \begin{tabular}[c]{@{}c@{}}Library  \vspace{-1mm} \\ Usage \end{tabular} & \begin{tabular}[c]{@{}c@{}}Issue \vspace{-1mm} \\  \end{tabular}    \\ \hline
 \multirow{2}{*}{$\mathbb{T}1$}     & \multirow{2}{*}{\footnotesize{$f(X)=X^TW_aX + W_bX$}} & ~\cite{cheung1991rotational}    & \multirow{2}{*}{$\mathcal{O}(n^2+n)$} & \multirow{2}{*}{$\mathcal{O}(n^2+n)$} &1-layer & $\times$   &\multirow{2}{*}{\scriptsize{\code{P2}\code{P3}\code{P4}}}   \\ \cline{3-3} \cline{6-6} \cline{7-7} 
  &  &~\cite{zoumpourlis2017non}   &  &   &1-layer &Torch & \\ \hline
\multirow{3}{*}{$\mathbb{T}1$}     & \multirow{3}{*}{$f(X)=X^TW_aX$}      &~\cite{redlapalli2003development}                   & \multirow{3}{*}{$\mathcal{O}(n^2)$} & \multirow{3}{*}{$\mathcal{O}(n^2)$} &1-layer& Matlab  &\multirow{2}{*}{\scriptsize{\code{P2}\code{P3}\code{P4}}}              \\ \cline{3-3} \cline{6-6} \cline{7-7} 
  &  &~\cite{jiang2019nonlinear}    &  &  &6-layer&TensorFlow & \\ \cline{3-3} \cline{6-6} \cline{7-7} 
  &  &~\cite{mantinicqnn}    &  & &4/18-layer &$\times$  & \\ \hline
 $\mathbb{T}2$ &$f(X)= W_aX^2$  &~\cite{goyal2020improved}    &$\mathcal{O}(2n)$  &$\mathcal{O}(n)$  &2-layer &$\times$ &\scriptsize{\code{P1}\code{P3}} \\ \hline
$\mathbb{T}3$  &$f(X)={(W_aX)}^2$  &~\cite{declaris1991novel}    &$\mathcal{O}(2n)$  &$\mathcal{O}(n)$ &1-layer &$\times$ &\scriptsize{\code{P1}\code{P3}} \\ \hline
$\mathbb{T}4$ &$f(X)=(W_aX) \circ (W_bX)$*  &~\cite{bu2021quadratic}   &$\mathcal{O}(3n)$  &$\mathcal{O}(2n)$  &5/10-layer &TensorFlow &\scriptsize{\code{P3}} \\ \hline
$\mathbb{T}1$\&2 &$f(X)=X^TW_aX +W_bX^2$  &~\cite{milenkovic1996annealing}    &$\mathcal{O}(n^2+2n)$  &$\mathcal{O}(n^2+n)$ &1-layer &$\times$  &\scriptsize{\code{P2}\code{P3}\code{P4}}\\ \hline
$\mathbb{T}2$\&4 &\footnotesize{$f(X)=(W_aX) \circ (W_bX)+W_cX^2$}  &~\cite{fan2018new}   &$\mathcal{O}(5n)$  &$\mathcal{O}(3n)$ &2-layer &Matlab &\scriptsize{\code{P3}} \\ \hline 
Ours &\footnotesize{$f(X)=(W_aX) \circ (W_bX)+W_cX$}  &This work   &$\mathcal{O}(4n)$  &$\mathcal{O}(3n)$ &Various &\textit{QuadraLib} &- \\ \hline 
\end{tabular}}
\leftline{\hspace{1mm}1. $X^T = \{x_1, x_2, ..., x_d\}$ is neuron's input and $f(X)$ represents neuron's output. For simplicity, we ignore bias $b$ here.} 
\leftline{\hspace{1mm}2. *:in~\cite{bu2021quadratic}, it only introduces the quadratic ResNet design. We list the design's general format.}
\leftline{\hspace{1mm}3. $\circ$ represents Hadamard product. $\dagger$: The second-order layer in the QDNN.}
% \XC{Don't mark with unknown, maybe we can say if they could implemented with an AI lib or not, and therefore highlight the necessity of our works. Also, types go first in the table.}
\label{table:2ndorderworks}
\vspace{-7mm}
\end{table*}
\normalsize

\vspace{-1mm}
\section{Drawbacks of the Existing QDNN Neuron Architecture Design}
\label{sec:prel}
\vspace{-1mm}

% \subsection{Characterization of Existing QDNNs}
% \vspace{-1mm}
% \emph{In this section, we investigate and categorize the existing QDNN work into four types according their n euron computation format.}

We first reviewed the current QDNN works in the last decades based on their corresponding key quadratic neuron computation formats and summarized them in Table.~\ref{table:2ndorderworks}.
According to the way that how the second-order term of input $X$ is introduced in each quadratic neuron, we can divide the current QDNN works into four types as shown in table: for $\mathbb{T}1$ design, each input $X$ with size $d$ will conduct out-product with $d\times d$ full-rank weight matrix\footnote{\scriptsize{The original format of the proposed quadratic neuron formulation in~\cite{cheung1991rotational, milenkovic1996annealing} is $\sum_{i=1}\sum_{j=1}W_{ij} x_i x_j$, which mathematically equals to $X^T W X$.}}; for $\mathbb{T}2$ design, the second-order term is realized by directly squaring each input $X$;
for $\mathbb{T}3$ design, the second-order term is coming from the square of a first-order neuron. Therefore, the size of weight parameter vector $W_a$ in both $\mathbb{T}2$ and $\mathbb{T}3$ keeps unchanged compared to the original first-order neuron; in $\mathbb{T}4$ quadratic neuron, the second-order term is calculated by the Hadamard Product of the two first-order neurons with different sets of weight parameters.

% \vspace{0.5mm}
% \subsection{Drawbacks Analysis}
% \vspace{0.5mm}

Most previous QDNN works focused on theoretically proving the stronger non-linearity and approximation ability of the single quadratic neuron, therefore they only applied QDNNs on some very simple learning tasks with small network structures, such as one fully connection layer for XOR Gate simulation~\cite{fan2018new, cheung1991rotational, redlapalli2003development}. 
However, by investigating the computation pattern of these existing QDNN works from the angle of practical usage, we identified several problems.

% \vspace{-0.5mm}
\noindent \textit{\textbf{\code{P1} Approximation Capability Issue:}} This issue is from $\mathbb{T}2$ and $\mathbb{T}3$ due to the insufficient trainable parameters.
% Approximation capability has impacts on the learning performance, which is usually indicated by the prediction accuracy. 
% All these QDNN designs demonstrate a better learning performance in their original papers. However, 
Specifically, as Table.~\ref{table:2ndorderworks} shows, compared to $\mathbb{T}1$ and $\mathbb{T}4$, $\mathbb{T}2$ and $\mathbb{T}3$ QDNN design only have one set of weight parameters while didn't introduce extra trainable parameters. According to~\cite{radosavovic2019network}, the approximation capability is highly related to the model capacity, namely, with a larger amount of trainable parameters, the model will show a higher learning performance potential. In that case, $\mathbb{T}2$ and $\mathbb{T}3$ model design will have lower theoretical learning performance (such issue will be further validated with numerical experiments in Section 5.2). 
\noindent \textit{\textbf{\code{P2} Computation Complexity Issue:}} This issue is from $\mathbb{T}1$
% \XC{Shall we specify the computation as neuron level?}
% Computation efficiency includes \textbf{space and time complexity}, which is determined by the model parameter size and computation workload. 
since it introduces a full-rank weight matrix in each neuron, which is a 4D array $W\in R^{C \ast r^4 \ast N \ast C}$ (where $C$, $N$, and $r$ represent the input channel number, filter number, and kernel size). Therefore, if we let $r^2$ equals to $n$, the time or space complexity of $\mathbb{T}1$ neuron is at least $\mathcal{O}(n^2)$. Such complexity will exponentially increase with a larger model structure regarding depth and width, easily introducing out-of-memory issue during model training. For example, in~\cite{mantinicqnn}, the parameter size of the original first-order \textit{ResNet} is only 0.2M while it dramatically increases to 128M for QDNN with same structure. 
% We will address such high computation complexity issue with our new neuron design in Section 3.

% Moreover, quadratic neuron will also introduce more intermediate parameters during training, which is not memory-efficient for the hardware (We will discuss it with more details and propose a corresponding solution in Section 3.3).

\vspace{-0.5mm}
\noindent \textit{\textbf{\code{P3} Converge Performance Issue:}} This issue is from $\mathbb{T}1$ to $\mathbb{T}4$ since we found that the second-order term in QDNNs will introduce critical gradient vanishing issue, impairing QDNN training convergence. 
% During training process, weight parameters in QDNNs will be updated via calculated gradients with certain gradient descent algorithm. However, we found that the second-order term in QDNNs will introduce critical gradient vanishing issue, impairing QDNN training effectiveness. 
Here, we use $\mathbb{T}4$ design $(W_aX)(W_bX)$ with plain network structure as an example. Assume $X^k$ represents the input of the $k^{th}$ layer. During back-propagation, the gradient for $X_k$ can be formulated as:
\vspace{-1.5mm}
\begin{equation}
\small
	\medmuskip=-2mu
	\thinmuskip=-2mu
	\thickmuskip=-2mu
	\hspace{-1mm}
	g(X^k) = \frac{\partial \ell}{\partial X^L} \cdot \frac{\partial X^L}{\partial X^k} = \frac{\partial \ell}{\partial X^L} \cdot \prod_{i={k+1}}^{L} X^i \prod_{i={k+1}}^{L} ((W_a^i)^2 + (W_b^i)^2)
	\label{eq:2ndtermissue}
	\vspace{-0.5mm}
\end{equation}
\normalsize
where $\ell$ indicates the loss value. We ignored activation function here since their gradients equal to one (when input larger than zero) during back-propagation. 
It is easily to find that: contrast to the traditional first-order DNN which only include weight parameters in their gradients, gradients in QDNNs contain both weights $\prod_{i={k+1}}^{L} ((W_a^i)^2 + (W_b^i)^2)$ and outputs from other layers $\prod_{i={k+1}}^{L} X^i$. 
    Since activation $X_i$ objects to norm distribution $X_i\sim N(0,1)$, $\prod_{i={k+1}}^{L} X^i$ is obviously much smaller than 1. 
Moreover, with depth increase, $\prod_{i={k+1}}^{L} X^i$ will gradually approach to zero, causing gradient vanishing. It should be noted that the same issue occurs for other kinds of second-order term design as well, which proves that it is the common issue for all QDNNs.

% \vspace{-0.5mm}
\noindent \textit{\textbf{\code{P4} Implementation Feasibility Issue:}} 
% This issue is from $\mathbb{T}1$.
Implementation feasibility indicates whether the designed QDNN is development-friendly with the current DNN libraries. As shown in Table.~\ref{table:2ndorderworks}, only a few of the existing QDNN works mentioned that they could be implemented on some DNN libraries, \textit{e.g.}, Matlab and Torch, while most of them didn't specify the capability of being supported by the current DNN libraries. Specifically, the currently DNN frameworks are all specifically designed for the first-order DNNs with formation $X^TW$, where $W$ is $n\times1$ when assuming the size of $X$ is $n\times1$. On the contrary, $\mathbb{T}1$ neuron has a bilinear formation $X^T W X$ where the size of $W$ is $n\times n$. Therefore, such neuron cannot be simply composed by two consecutive linear neurons. We need to rewrite the entire convolution operation to add extra multiplication between $W$ and the second $X$. Therefore, such type of neurons are not implementing-friendly.

% only a few of the existing QDNN works mentioned that they could be implemented on some DNN libraries, \textit{e.g.}, Matlab and Torch, while most of them didn't demonstrate the capability of being supported by the current DNN libraries.
% Specifically, the currently DNN frameworks are all specifically designed for the first-order DNNs with formation $X^TW$. However, $\mathbb{T}1$ neuron is the special bilinear version ($A^T W B$, when $A=B=X$). We need to rewrite the entire convolution operation to add extra multiplication between $W$ and the second $X$. Therefore, such type of neurons are not implementing-friendly. 

\vspace{0.5mm}
\noindent \textit{\textbf{\code{P5} Structure Design Issue:}}
The existing QDNN works suffer from two sub-problems in terms of QDNN structure design: first, since they propose distinct quadratic neuron architectures and most of are not capable for the current DNN libraries, it is difficult to reproduce these QDNN design; second, most of them merely leverage shallow model structures \textit{e.g.}, one or two layers, to verify the designed neuron capability, while such shallow model structures are not practicable for common-used learning applications such as image classification or object detection.  However, identifying an optimal network structure for a given learning scenario usually needs to introduce significant design efforts, such as Network Architecture Search (NAS) in the first-order DNN design.

\vspace{-1mm}
\noindent \textit{\textbf{\code{P6} Memory Usage Issue:}}
As Table.~\ref{table:2ndorderworks} shows, some existing QDNN design such as~\cite{cheung1991rotational, micikevicius2017mixed, fan2018new} introduce more intermediate parameters during training. During training process, QDNNs needs to be executed forwardly to get loss and other intermediate parameters such as activation. These intermediate parameters needs to be cached in computing unit's memory (\textit{e.g.} GPU) until finishing gradients computing during backward, thereby will introduce high memory cost, eventually causing memory inefficient-usage problem.   
% some existing QDNN design such as~\cite{cheung1991rotational, micikevicius2017mixed, fan2018new} introduce more weight parameters inside each neuron. Although increased approximation capability, they also introduce more intermediate parameters during training.
% This is because during training process, QDNNs needs to be executed forwardly to get loss and other intermediate parameters that are necessary to compute the gradients of the model parameters, such as activation. These intermediate parameters needs to be cached in computing unit's memory (\textit{e.g.} GPU) until finishing gradients computing during backward, thereby will introduce high memory cost, eventually causing memory inefficient-usage problem. 
% In order to solve this issue, we optimize the training scheme in our proposed QDNN library, which will be discussed in Section 4.3. 

% Moreover, quadratic neuron will also introduce more intermediate parameters during training, which is not memory-efficient for the hardware (We will discuss it with more details and propose a corresponding solution in Section 3.3).

In the rest of this paper, we addressed the above six problems from two main perspectives: 
\begin{itemize}
\vspace{-3mm}
    \item From theoretical \textbf{neuron architecture optimization perspective}, we proposed a new quadratic neuron design to solve problems \code{\textit{P1}} to \code{\textit{P4}} (Section 3).
    % low approximation capability, intensive computation, hard converge and infeasible implementation (\code{\textit{P1}} to \code{\textit{P4}} ) (Section 3).
    \vspace{-1mm}
    \item From practical \textbf{model construct and training optimization perspective}, we proposed \textit{QuadraLib}, which can generate optimal QDNN model structure for the practical deployment (\code{\textit{P5}} ) and address the inefficient memory usage problem  (\code{\textit{P6}} ) (Section 4). 
    % \vspace{-3mm}
    % \item From \textbf{model training perspective}, we optimized back-propagation scheme in \textit{QuadraLib} to address the inefficient memory usage problem during training phase (\code{\textit{P6}} ) in Section 5. 
\end{itemize}
\vspace{-4mm}
% \begin{itemize}
% \vspace{-4mm}
%     \item From \textbf{neuron design perspective}, we some existing QDNN design suffer from low approximation capability, intensive computation, hard converge and infeasible implementation (\code{P1} to \code{P4} ).
%     \vspace{-3mm}
%     \item From \textbf{model construct perspective}, the existing QDNN work didn't provide complex QDNN model structure for the practical deployment (\code{P5} ). 
%     \vspace{-3mm}
%     \item From \textbf{model operation perspective}, current QDNN work will introduce inefficient memory usage problem during training phase (\code{P6} ). 
% \end{itemize}
% \vspace{-4mm}

\vspace{-0.5mm}
\section{QDNN Neuron Architecture Optimization}
\label{sec:prel}
\vspace{-1mm}

\begin{figure}[t]
	\centering
	\captionsetup{justification=centering}
	\vspace{-1.5mm}
	\includegraphics[width=3.2in]{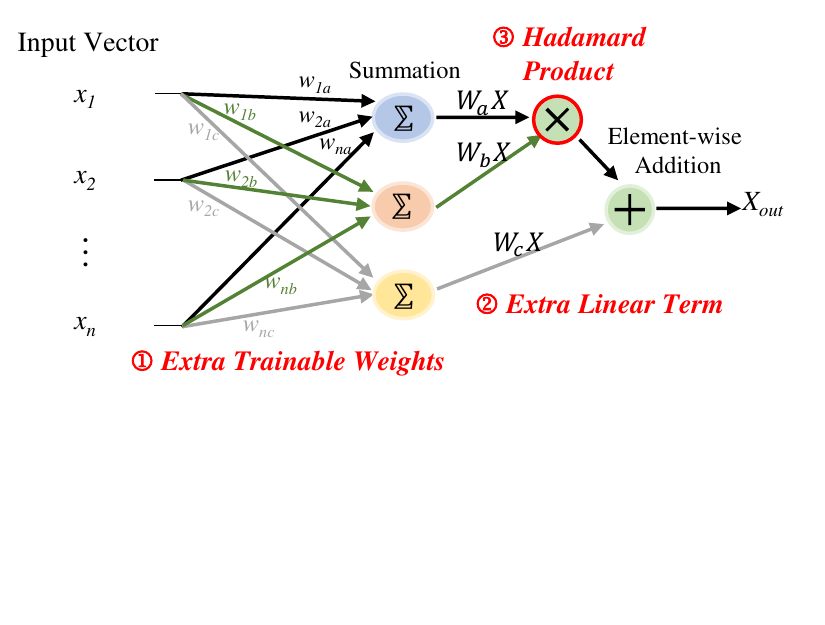}
	\vspace{-28mm}
	\caption{The New Quadratic Neuron Design\vspace{3mm}}
	\vspace{-7mm}
	\label{quatratic_neuron}
\end{figure}

\vspace{0mm}
\subsection{New Neuron Architecture Design}
\vspace{0mm}

Through the above analysis, we identified that the existing QDNN works show one or more drawbacks due to their neuron computation format design. 
Fortunately, from the above performance analysis, we can obtain several \textit{\textbf{design insights:}} 
1) The second-order term in QDNN is expected to introduce extra trainable parameters instead of just using the original one set of weight parameters in the first-order neuron. 
2) Besides second-order term, the new quadratic neuron should include other term to prevent the gradient vanishing issue. 
3) Since the original out-product between weight matrix and input vector introduces intensive computation workload as well as memory cost, replacing it with Hardmard product can significantly improve designed neurons' computation efficiency. 
4) To achieve similar implementation feasibility as the first-order DNNs, the proposed quadratic neuron should be easily assembled via multiple first-order neurons.

% In order to increase the approximation capability, the quadratic neuron is expected to introduce extra trainable parameters instead of just using the original one set of weight parameters in the linear neuron. 
% 2) The original out-product between weight matrix and input vector introduce intensive computation workload as well as memory cost. On the contrary, using Hardmard product to replace the heavy out-product operation will significantly improve designed neurons' computation efficiency. 
% % 3) To decrease training difficulty via introduced extreme values (speedup training convergence), the designed neuron should avoiding directly squaring input.
% 3) To achieve similar implementation feasibility as the 1$^{st}$ order CNN, the quadratic neurons are expected to be easily assembled by multiple linear neurons.

Based on the above design insights, we optimize the second-order term and introduce new linear term into quadratic neuron design, and propose a new quadratic neuron format (as shown in Fig.~\ref{quatratic_neuron}), which can comprehensively address the identified neuron design issue in the current QDNN works. The proposed new quadratic neuron is formulated as: 
\vspace{-0mm}
\begin{equation}
\small
	\medmuskip=-0mu
	\thinmuskip=-0mu
	\thickmuskip=0mu
	\hspace{-1mm}
	f(X) = (W_a X)\circ(W_b X) + (W_cX).
	\label{eq:2ndneuron3}
    \vspace{-0.5mm}
\end{equation}
\normalsize
It should be noticed here, our proposed neuron design is similar to ~\cite{fan2018new}. 
~\cite{fan2018new} mainly focus on improving the approximation capability by introducing two types of second-order term. However, as discussed in \textit{\code{P3}}, such design will amplify gradient vanishing issue, hurting its converge performance. On the contrary, we address this issue by replacing $W_cX^2$ with a linear term $W_cX$. 
% compared to~\cite{fan2018new}, our design replaces $W_cX^2$ with a linear term $W_cX$. \ZX{208 difference issue,ignore some issue}
% which can improve approximation capability and training effectiveness (will be further discussed below). 
We will further analyze the superiority of our design in terms of approximation capability, converge performance, computation complexity, and implementation feasibility.

\subsection{Theoretical Performance Analysis}

\noindent \textit{\textbf{Extra Weights and Linear Term for Approximation Capability (\code{P1} ) Improvement:}}
The approximation capability improvement of our design is achieved from two aspects: first, similar to $\mathbb{T}4$ design, our design introduces two sets of weight to form the second-order term, providing extra trainable parameters (shown as $\raisebox{.5pt}{\textcircled{\raisebox{-.9pt} {1}}}$ in Fig.~\ref{quatratic_neuron}). Second, we add an extra linear term in our design (shown as $\raisebox{.5pt}{\textcircled{\raisebox{-.9pt} {2}}}$ in the figure). Specifically, let's use a $L$-layer QDNN with plain structure (e.g. \textit{VGG-16}) as an example. Such model can be formulated without/with linear term $W_c X^{k}$ as:
\vspace{-5.4mm}

\begin{equation}
\small
	\medmuskip=-0mu
	\thinmuskip=-0mu
	\thickmuskip=0mu
	\hspace{-1mm}
	Y=\prod \limits_{l=1}^L W_a^{l}W_b^{l} X^{2^L}  \stackrel{}{\longrightarrow} \ \ Y=\sum_{i=1}^{2^L} \prod \limits_{l=1}^L W_a^{l}W_b^{l}W_c^{l} X^{i},
	\label{eq:ep1_1}
	\vspace{-2.2mm}
\end{equation}
\normalsize
where $\prod\limits_{l=1}^L W_a^{l}W_b^{l}$ represents the product over a set of weight $W_a^l$ and $W_b^l$. The first high-order polynomial only includes a $2^L$ order term regarding input $X$. According to Chebyshev approximation~\cite{hernandez2001chebyshev}, the approximation error introduced by such polynomial is approached to a $X^{2^L -1}$ term. On the contrary, the second polynomial includes multiple terms with $X$ from the first-order to $2^L$ order, which means its approximation error is approached to zero. Therefore,  by adding linear term, our neuron design can significantly improve approximation capability.

\vspace{0.5mm}
\noindent \textit{\textbf{Hadamard Product for Computation Complexity  (\code{P2} ) Optimization:}}
In order to avoid huge computing workload and memory coast of $\mathbb{T}1$ design, the original out-product between weight matrix and input vector is replaced by a Hadamard product ($\raisebox{.5pt}{\textcircled{\raisebox{-.9pt} {3}}}$ in figure), which will significantly improve designed neurons' computation efficiency from $\mathcal{O}(n^2)$ to $\mathcal{O}(n)$ regarding time/space complexity. In Section 3.3, we will further discuss how we boost computation efficiency for QDNN during training through removing unnecessary intermediate parameters.

\vspace{0.5mm}
\noindent \textit{\textbf{Linear Term for Converge Performance (\code{P3} ) Enhancement:}}
As discussed in Section 2.2, all kinds of second-order terms will introduce critical gradient vanishing issue, decreasing training effectiveness. On the contrary, we identified that the added first-order term ($\raisebox{.5pt}{\textcircled{\raisebox{-.9pt} {2}}}$) in our neuron design can solve this issue. In the first-order DNN model, we leverage identity mapping ($h(X^{k-1})$~\cite{he2016identity}) to provide an extra term $\frac{\partial \ell}{\partial X^L}$
(the gradient of loss to the last layer) during each layer's gradient calculation. Such extra term ensures that information can propagate directly to each shallow layers.
We found that the first-order term $W^c x^k$ in our design not only can improve neuron approximation capability, but also can work as the identity mapping to solve the gradient vanishing issue. 
Specifically, the quadratic layer with our design neuron is: $X^{k+1} = (W_a^k X^k W_b^k X^k) + W_c^k X^k$. During back-propagation, the gradient of $X^k$ is:
\vspace{0.5mm}
\begin{equation}
\small
	\medmuskip=-0mu
	\thinmuskip=-0mu
	\thickmuskip=0mu
	\hspace{0.5mm}
	\frac{\partial X^{k+1}}{\partial X^k} = X^k(W_a^2 + W_b^2) + W_c, 
	\label{eq:type3}
\end{equation}
\normalsize
where the additional term $W_c$ is the main contribution for the gradient calculation in the original first-order DNNs. It can cooperate with Batch Normalization layer and activation function (\textit{e.g.} ReLU) to prevent gradient vanishing issue\footnote{\scriptsize{It may still meet model degeneration issue~\cite{orhan2017skip} for deep plain network structure, which can be further solved by using our model construction method or \textit{ResNet} structure.}}.

\vspace{0.5mm}
\noindent \textit{\textbf{First-order Neuron Combination for Implementation Feasibility (\code{P4} ) Improvement:}}
In order to have high implementation feasibility, our designed quadratic neuron is the combination of three traditional first-order neurons (shown in Fig.~\ref{quatratic_neuron}): the first two conduct Hadamard Product and then the product will sum with the third one. Different with $\mathbb{T}1$ design that need to create the customization convolution operation, all the operations defined in our neuron are supported by the current DNN libraries, without introducing extra implementation efforts. Moreover, these operations already be optimized by the DNN libraries from compiler-level, guaranteeing an optimal computing performance.

% \vspace{-1mm}
\section{QuadraLib for QDNN Design Exploration}
\label{sec:prof}
\vspace{0mm}

In the last section, we proposed a new quadratic neuron design to address previous drawbacks (\code{\textit{P1}} to \code{\textit{P4}} ) from theoretical aspect.
According to library capability analysis in the last column of Table.~\ref{table:2ndorderworks}, there lacks a library to support the implementation of all kinds of the existing QDNNs. 
To further address (\code{\textit{P5}}\code{\textit{P6}} ) and conduct in-depth numerical analysis as well as achieve design exploration on QDNNs, we proposed \textit{QuadraLib}, \textit{an extended PyTorch-based Python library to provide a set of complementary components for QDNNs.} 
% \textit{\textbf{Therefore, in order to conduct in-depth numerical analysis and achieve further exploration on QDNNs, a library specifically designed for QDNNs is highly desired. }}
% Meanwhile, \textit{QuadraLib} can also address the structure complexity issue (\code{\textit{P5}} )and inefficient memory usage issue (\code{\textit{P6}} ) during QDNN training efficiency.

\begin{figure*}[!t]
	\centering
	\captionsetup{justification=centering}
	\vspace{2mm}
 	\includegraphics[width=6.6in]{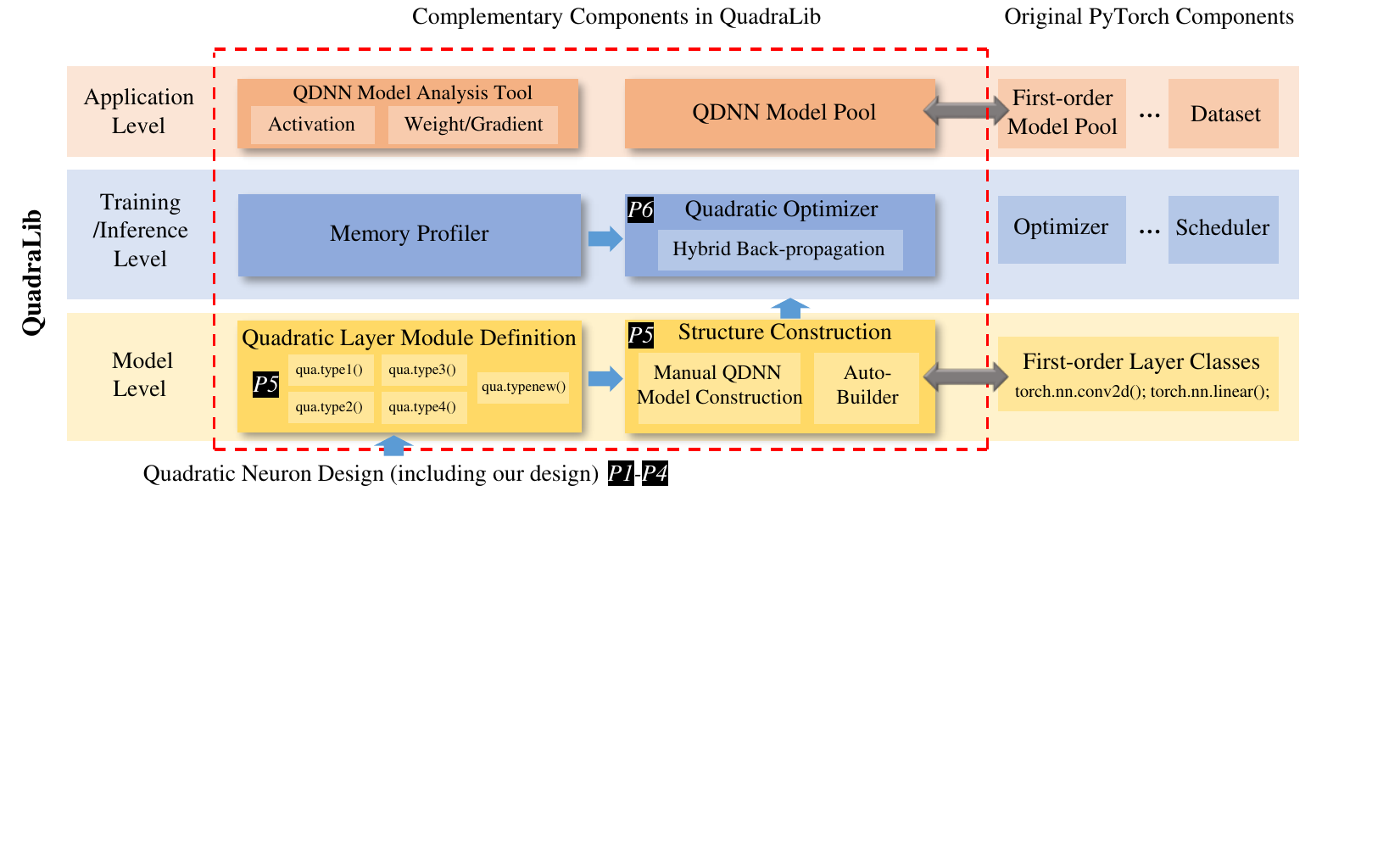}
	\vspace{-46mm}
	\caption{\textit{QuadraLib} Architecture Design}
	\vspace{-5mm}
	\label{architecture}
\end{figure*}

\vspace{0mm}
\subsection{Library Overview}
% \vspace{-1mm}

% Section 2 and Section 3 provides analysis of different QDNN neuron types and our QDNN architecture optimization. 
% To further support the QDNN design exploration, we proposed \textit{QuadraLib}, \textit{an extend Python library built on PyTorch to provide a set of complementary components for QDNNs.} 
Fig.~\ref{architecture} shows our \textit{QuadraLib}'s design overview: 
\begin{itemize}
\vspace{-3mm}
    % \item In Model Level, \textit{QuadraLib} defines a set of \textit{encapsulated layer modules} including all existing quadratic neuron types to facilitate the QDNN building and exploration. Based on that, \textit{QuadraLib} also provides a \textit{ converter} that can convert any first-order candidate model structure to a corresponding QDNN with optimal performance (Section 4.2).
    \item In Model Level, \textit{QuadraLib} defines a set of \textit{encapsulated layer modules} to support all existing quadratic neuron types. Based on that, \textit{QuadraLib} provides QDNN model construction flow and design insights to guarantee a decent accuracy performance of the built QDNN model. Furthermore, an auto-builder is proposed to extend QDNN design by converting any first-order model to a corresponding QDNN version.
\vspace{-2mm}
\item In Training/Inference Level, \textit{QuadraLib} leverages a memory profiler\footnote{\scriptsize{Currently, we use default memory profiler in Pytorch.}} to monitor the memory cost of the generated QDNN model during the training phase. A \textit{Quadratic Optimizer} is developed that applied a \textit{hybrid back-propagation scheme} to QDNN model training, which further enhances the training efficiency of QDNNs (Section 4.3).
\vspace{-2mm}
\item In Application Level, \textit{QuadraLib} also provides model analysis tools such as \textit{activation and weight/gradient distribution visualization}. Leveraging such visualization tools, we reveal the distinct feature extraction capability of QDNN vs. first-order DNNs (Section 5).

\end{itemize}

% \begin{itemize}

% \vspace{-4mm}
% \item On the algorithm level, we already analyzed the existing QDNN design in Section 2 and proposed our new quadratic neuron design in Section 3. They can work as the mathematical foundation for \textit{QuadraLib}.

% \vspace{-2.5mm}
% \item On the model level, we created the second-order layer class to optimize QDNN implementation feasibility so that users can conveniently build QDNN models and conduct comprehensively numerical analysis. 

% \vspace{-2.5mm}
% \item On the operator-function level, we addressed the computation efficiency issue by providing a hybrid back-propagation scheme, thereby the users can efficiently explore applications with large computation cost. 

% \vspace{-2.5mm}
% % \item For the Top-level APIs, we provide two \textit{QuadraLib} tools to address the structure practicality issue and help user have in-depth analysis for QDNN models, which will be discussed in Section 4.
% \item On the Top-level APIs, we first provided a QDNN model converter, which can automatically convert the existing DNN models to QDNN format with optimal structure and performance.

% \vspace{-2.5mm}
% \item We also provided model analysis tools and demonstrated how we leverage these tools to analyze QDNN activaiton and weight/gradient in Section 6.

% \end{itemize}

\vspace{-1.5mm}
\subsection{QDNN Model Construction}

As aforementioned, the existing QDNN works fail to provide a unified neuron design, QDNN structure design, as well as implementations (\code{P5} ). \textit{QuadraLib} addresses this issue by providing a series of QDNN model construction help: first, it defines the encapsulated quadratic layer modules as the fundamental units of model construction. Next, we can either use the quadratic layer module to construct a full QDNN model from scratch or combine it with other first-order layer modules in the pre-defined construction functions. Furthermore, we proposed an auto-builder which leverages knowledge of the existing first-order DNN to extend the QDNN model structure design.

% Consider the construction efficiency and convenience, we further proposed a QDNN auto-builder to convert an existing first-order DNN to a corresponding QDNN version with optimal performance.

% to enable the manual QDNN construction flow  

% including the encapsulated quadratic layer module, QDNN model construction flow, and QDNN model auto-builder. 

% \ZX{first define module, then simply replacing, for convinnent use, we can further converter}.
% Similar to the first-order DNN building in PyTorch, QDNN structure construction is assembled by various layer modules (\textit{e.g.} quadratic layer, pooling layer, batch normalization layer, etc.). Therefore, the most effective approach to solve the model complexity issue is to first cooperate with the layer modules in model level, which introduce two specific design challenges: 1) how to define an efficient quadratic layer module; 2) how to generate the optimal QDNN with minimum efforts. 
% Therefore, as shown in Fig.~\ref{architecture}, we first define the quadratic layer module to efficiently provide the fundamental unit for QDNN construction and further propose a model structure builder to automatically convert a given existing first-order DNN model to it QDNN version with optimal performance. 

\noindent \textit{\textbf{Encapsulated Quadratic Layer Module:}}
% When we build the library, we identified 
% \XC{Based on the analysis above, we identified that, the current design still suffer from xxx issue.
% When we build the library, the most effective approach is to incorporate the xx layer xx modelue. liangcheng
% And we will desolve the implementation challenges a sxxxxxxxx. sicheng
% Specificially, the major challenges falls into xxx and we can leverage xxxx. sicheng.}
% Based on the drawback analysis in Section 2, we demonstrated that most existing QDNN work failed to provide a complex QDNN structure with optimal performance during deployment (\code{P5} ). 
Similar to construct the first-order DNNs in PyTorch, QDNN construction is expected to be assembled by various quadratic layer modules. Therefore, \textit{QuadraLib} first defines and encapsulates the quadratic layer modules as fundamental units to alleviate the manual QDNN construction complexity.

% In PyTorch, models are created with basic components called modules. Each module is equivalent to a layer in DNN.
% Users can construct their own models like assembling building blocks.
% In this part, we are aiming to optimize QDNN implementation feasibility via extending such assembling concept. 

% When designing the quadratic layer via the default first-order convolutional layer in PyTorch, there are two critical issues: 
% 1) we have to manually initialize every first-order layer and add them into computational graph. Such operation is tedious and code-inefficient, and will become more serious with the increasing of model depth.
% 2) most DNN models (\textit{e.g.} \textit{VGG-16} or \textit{ResNet-56}) in PyTorch use loops to improve code-efficiency by adding each layer to model structure. 
%     However, if we want to build QDNNs based on these loop functions, it will be hard to directly replace the original single first-order layer module with a combination of multiple linear layer modules.

% In order to solve the above two challenges, we abstract quadratic layer as a new layer module in \textit{QuadraLib}.
Specifically, various quadratic layers including our proposed one are written as new $nn.module$ separately. Once creating a QDNN, the desired quadratic layer can be introduced as $qua.type\#()$ (here,$\#$ represents different types of quadratic layer). 
Since our layer module is inherited from PyTorch, it supports flexible layer parameters setting such as kernel-size, padding/stride, depth-wise/point-wise.
By doing that, the customized quadratic layer and the original first-order layer share the same instantiation and computation graph definition methodology, thereby can easily replace any original first-order layer module in a given model to create a QDNN version.

% \begin{figure}[!t]
% 	\centering
% 	\captionsetup{justification=centering}
% 	\vspace{0.5mm}
% 	\includegraphics[width=3.3in]{model_layer}
% 	\vspace{-31.5mm}
% 	\caption{A QDNN Structure with Two Quadratic Layers}
% 	\vspace{-6.5mm}
% 	\label{model_layer}
% \end{figure}

\noindent \textit{\textbf{Manual QDNN Model Construction:}}
Since our proposed quadratic layer modules share the same characteristics with the ordinary first-order layer modules in PyTorch, the model construction process in \textit{QuadraLib} is aligned with PyTorch, which is user-friendly. Specifically, during model construction, we first created a structure configuration file. By defining the model structure parameters such as depth and width, such configuration file can guide the structure design. Here are some insights for QDNN model design during configuration file definition: 1) since quadratic neuron has higher capability, the depth of QDNN can be reduced, thereby decrease the model computation cost and also avoids the potential gradient vanishing or model degeneration issues discussed before; 2) since second-order term will generate extreme values, batch-normalization layer is significantly important for QDNN to regulate the output activation values; 3) QDNNs with small network structures don't need activation functions (\textit{e.g.} ReLU) due to the high capability of quadratic neuron. However, when QDNN depth increases, activation functions are important since they can prevent gradient vanishing as discussed in Section 3.2. Secondly, once we determined configuration file, we will insert quadratic layer modules into a construction function and the entire QDNN model can be defined as a layer sequence. Here we demonstrated a code piece example  of our construction function with \textit{VGGNet} structure in the following figure.
\begin{lstlisting}
// A code piece example of our construction function.
import QuadraNeuron as quad
def model_construct(self, cfg):
    ...
	for v in cfg:
		layers += [qua.type1(in_channels, v), nn.ReLU()]
		in_channels = v
	return nn.Sequential(layers)
\end{lstlisting}
\vspace{-2.5mm}
It is easily to see that our quadratic layer module can directly inserted into construction function with other layer types (\textit{e.g.} $nn.ReLU()$), which is flexible and convenient. 
At last, users can easily import the construction function into their code and  instantiate a corresponding QDNN model. 

\vspace{-0.3mm}
\noindent \textit{\textbf{QDNN Auto-Builder:}}
For an existing learning task, manually designing a QDNN model from scratch needs a lot of prior domain experience and can involve significant effort, such as detector backbones.
    Therefore, besides to manually construct QDNN from scratch, another effective approach for efficient QDNN construction is to leverage the existing first-order DNN model pools, which already include various sophisticated pre-defined first-order DNN structure for different learning tasks. 
Therefore, \textit{QuadraLib} provides a QDNN auto-builder to convert the first order DNN to a corresponding QDNN without designing from scratch.
    
% As Fig.~\ref{architecture} shows, the proposed structure builder includes three sub-components: capacity estimator, model selector, and converter. Capacity estimator will provide the desired model computation level in terms of parameter size and computation workload. Model selector further selects the potential first-order DNN from first-order DNN pool according to the estimated model computation level. After that, the converter will convert the first-order DNN to a corresponding suitable QDNN version.

% 1. 从第一层到最后一层，逐层替换，最后变成一个hybrid的网络
% 2. 开始training一定epochs，做profiling
% 3. 利用公式确定哪些层可以去掉。
% 4. 最后重新接着train 

To guarantee the converted QDNN model has optimal performance (\textit{e.g.} prediction accuracy and model depth), we conducted two operations: layer replacement and heuristic-based layer reduction. Specifically, we first iteratively replaced the first-order layer module in the original construction function with our quadratic layer module from shallow layer to deep layer. Secondly, we identified which layers in the converted model can be removed to achieve a more compacted model structure.
    In order to achieve that, we used layer performance indicator from~\cite{xu2019reform}, which can be defined as:
\begin{equation}
\vspace{0mm}
\small
	\medmuskip=-0mu
	\thinmuskip=-0mu
	\thickmuskip=0mu
	\hspace{0mm}
	RI = \frac{P(M_{par}) P(T_{lat})}{\vartriangle Acc}, 
	\label{eq:remove}
\end{equation}
\normalsize
where $P(M_{par})$ and $P(T_{lat})$ represent the ratio of the target layer to the entire model in terms of parameter amount and computation workload. $\vartriangle Acc$ represents the accuracy drop when removing the target layer. 
The layer with high parameter size and computation workload while low accuracy contribution will have a high $RI$ value, which means that it can be removed firstly during our heuristic-based layer reduction. Based on the above two operations, auto-builder can finalize the optimal model structure for QDNN,  providing extra model structure sources for QDNN model pool in Application Level of \textit{QuadraLib}.

\begin{figure}[t]
	\centering
	\captionsetup{justification=centering}
	\vspace{0mm}
	\includegraphics[width=3.3in]{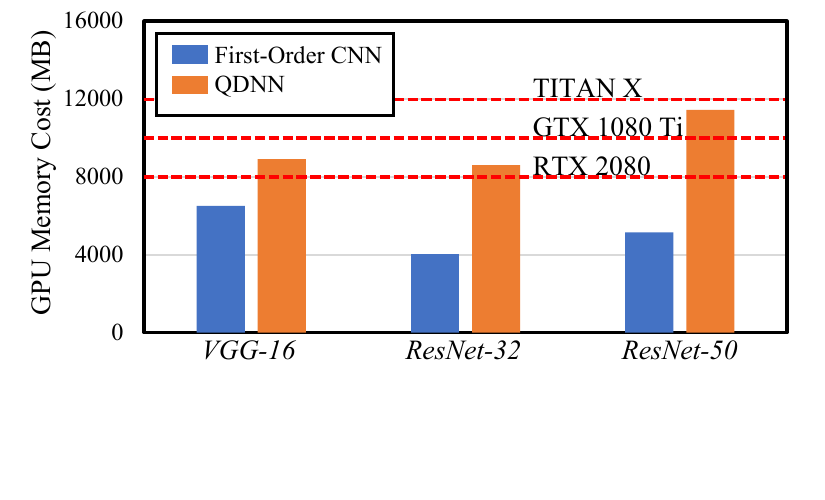}
	\vspace{-21.6mm}
	\caption{GPU Memory Cost Evaluation}
	\vspace{-5.94mm}
	\label{memory}
\end{figure}

\vspace{-0.2mm}
\subsection{QuadraLib Training Optimization}
\vspace{-0.2mm}

Based on the analysis in Section 2, we identified the current QDNNs still suffer from inefficient memory usage issue ( \code{P6} ) since they will generate more intermediate parameters during training phase. \textit{QuadraLib} address this problem in Training/Inference Level via proposing a quadratic optimizer. Specifically, the model will be first evaluated by a \textit{memory profiler} to exam whether it is under potential out-of-memory risk. 
Fig.~\ref{memory} shows the GPU memory usage of the first-order DNNs and $\mathbb{T}2$\&4 QDNN~\cite{fan2018new} monitored by our \textit{memory profiler}. The two evaluated models share the same network structure and batch size is set as 512. 
It is clearly to see that the first-order DNN training can be satisfied by most widely-used GPUs while QDNNs with same layers will introduce more memory cost and may cause out-of-memory issue during training phase. Once \textit{memory profiler} identified inefficiency memory usage issue, \textit{QuadraLib} will optimize the training via a proposed hybrid back-propagation scheme.

% When we build QDNN library, the most effective approach to solve such problem is to optimize the training scheme in Operation function level. 

% \XC{Based on the analysis above, we identified that, the current design still suffer from xxx issue.
% When we build the library, the most effective approach is to incorporate the xx layer xx modelue. liangcheng
% And we will desolve the implementation challenges a sxxxxxxxx. sicheng
% Specificially, the major challenges falls into xxx and we can leverage xxxx. sicheng.}

% As we discussed in the computation efficiency issue in Section 2.2, QDNNs introduce inefficient memory usage since they will generate more intermediate parameters during training phase. Specifically, QDNN needs to be executed forwardly to get loss and other intermediate parameters that are necessary to compute the gradients of the model weights, such as activation. These intermediate parameters needs to be stored in computing unit's memory (\textit{e.g.} GPU) until finishing gradients computing during backward. Fig.~\ref{memory} shows the GPU memory usage of the first-order DNNs and Type-4 QDNN~\cite{fan2018new} with same network structure when batch size is setting as 512. 
% It is clearly to see that the first-order DNN training can be satisfied by most widely-used GPUs while QDNNs with same layers will generate much more memory cost and may cause out-of-memory issue during training phase.

\begin{figure}[t]
	\centering
	\captionsetup{justification=centering}
	\vspace{0mm}
	\includegraphics[width=3.3in]{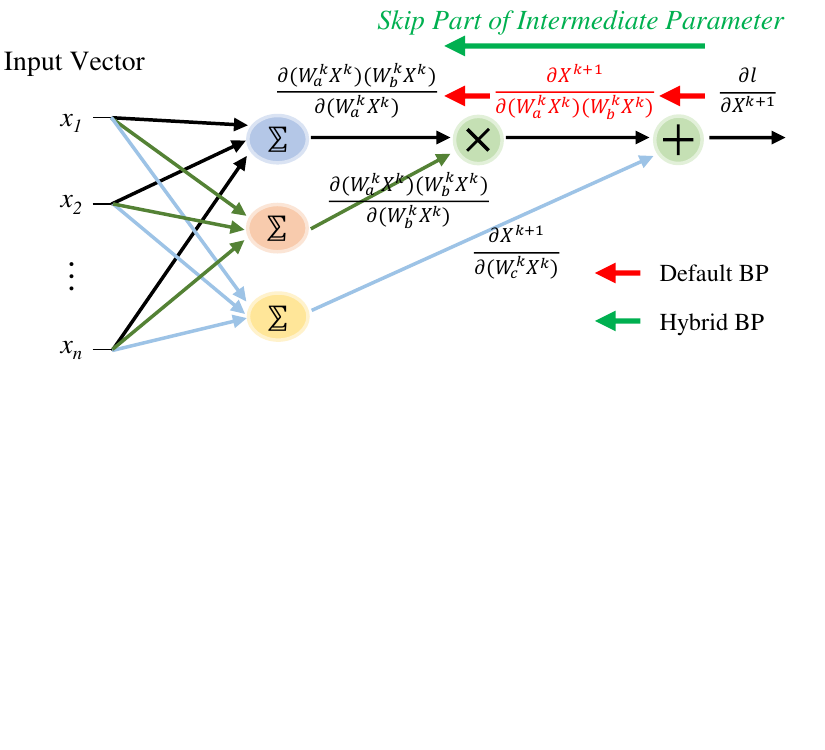}
	\vspace{-47mm}
	\caption{Back-propagation in QDNN Training}
	\vspace{-6mm}
	\label{customized_bp1}
\end{figure}

\noindent \textit{\textbf{Hybrid Back-propagation for Memory Efficiency:}}
In the most DNN libraries, gradient calculation during back-propagation is supported by Auto-Differentiation (AD) with reverse-mode ~\cite{baydin2018automatic}, which means the partial derivatives will backwardly go through each layer from output to input. 
The red arrow in Fig.~\ref{customized_bp1} shows the default back-propagation process for our quadratic layer. Assume we want to update $W_a^k$, its gradient can be calculated according to the chain rule as:
\vspace{-2mm}
\begin{equation}
\small
	\medmuskip=-2mu
	\thinmuskip=-2mu
	\thickmuskip=-2mu
	\hspace{-1mm}
	\frac{\partial{\ell}}{\partial{W_a^{k}}} = \frac{\partial \ell}{ \partial X^{k+1}} \cdot \frac{\partial X^{k+1}}{\partial (W_a^{k}X^k)(W_b^k X^k)}\cdot \frac{\partial (W_a^{k}X^k)(W_b^k X^k) }{\partial (W_a^{k}X^k)} \cdot \frac{\partial (W_a^{k}X^k)}{\partial W_a^k}. 
	\label{eq:AD}
\end{equation}
\normalsize
During this process, all the intermediate parameters needs to be stored on GPU memory. 
Compared to AD, Symbolic Differentiation (SD) can derive the symbolic expression of each parameter's partial gradients before the back-propagation. Therefore, training process doesn't need to save every intermediate parameters on the memory. For example, Eq.~\ref{eq:AD} can be simplified as: 
\vspace{-0mm}
\begin{equation}
\small
	\medmuskip=0mu
	\thinmuskip=0mu
	\thickmuskip=0mu
	\hspace{-1mm}
	\frac{\partial{\ell}}{\partial{W_a^{k}}} = \frac{\partial \ell}{ \partial X^{k+1}} \cdot (W_b^k X^k)\cdot X^k,
	\label{eq:SD}
\vspace{-1mm}
\end{equation}
\normalsize
which means that we don't need intermediate term $\frac{\partial X^{k+1}}{\partial (W_a^{k}X^k)(W_b^k X^k)}$ during SD process.
Therefore, we proposed a hybrid back-propagation scheme to decrease the memory usage during QDNN training, which is the combination of AD and SD. Indicated by green arrow in Fig.~\ref{customized_bp1}, for each quadratic layer, when obtaining input values and output values after forward process, we use SD to calculate the corresponding gradients of weight parameters and input during back-propagation. Therefore, only necessary intermediate parameters will be stored before the backward. 
On the other hand, since the gradients of other layers such as batch normalization are not easy to formulate in SD, we can still leverage AD to calculate their values. 

\noindent \textit{\textbf{Quadratic Optimizer Implementation:}}
% \XC{Based on the analysis above, we identified that, the current design still suffer from xxx issue.
% When we build the library, the most effective approach is to incorporate the xx layer xx modelue. liangcheng
% And we will desolve the implementation challenges a sxxxxxxxx. sicheng
% Specificially, the major challenges falls into xxx and we can leverage xxxx. sicheng.}
In the original PyTorch library, we can simply leverage $\textit{model.backward()}$ to conduct AD with gradient calculation. In order to achieve our hybrid back-propagation, we need to modify the backward computing flow inside each quadratic layer with SD. 
Therefore, we provide another version of our quadratic layer. Instead of be defined as $\textit{nn.module}$ which can only rely the default AD, we build it by inheriting $\textit{torch.autograd.Function}$ so that we can manually define both the forward and backward computing process inside the quadratic layer as individual $\textit{def forward}$ and $\textit{def backward}$. 
% Specifically, in the $\textit{def backward}$, we can give the symbolic expression of each weight parameter without calling extra intermediate parameters. 
Meanwhile, in $\textit{def forward}$, we use gradient checkpointing ($\textit{torch.utils.checkpoint.checkpoint}$) to make sure quadratic layer will not store these intermediate parameters after the forward process.

Besides model construction and training optimization, \textit{QuadraLib} also provides other components in Application Level. \textit{e.g.} model analysis tools (discussed in Section 5.)

\vspace{-1mm}
\section{Experimental Evaluation}
\label{sec:expe}

\begin{table}[!t]
\small
\renewcommand\arraystretch{1}
\centering
\vspace{-2mm}
\caption{Performance Evaluation of Various Quadratic Neuron Designs with Deep Network Structures.}
\vspace{2mm}
\setlength{\tabcolsep}{1.5mm}{
\begin{tabular}{@{}ccccccc@{}}
\toprule
       & \multicolumn{2}{c}{VGG-8} & \multicolumn{2}{c}{VGG-16} & \multicolumn{2}{c}{ResNet-32}  \\
       & Train          & Test       & Train         & Test        & Train          & Test          \\ \midrule
$\mathbb{T}2$ & 91\% &81\%   & \textcolor{red}{10\%}  &\textcolor{red}{10\%}  &99\% &89\%   \\
$\mathbb{T}3$ & 93\%  &85\%    & \textcolor{red}{10\%}   &\textcolor{red}{10\%}   &99\%  &89\% \\
$\mathbb{T}4$ & 94\%  &85\%      &\textcolor{red}{10\%}  &\textcolor{red}{10\%}   &99\% &89\%   \\
$\mathbb{T}4$$+$Identity   & 95\%   & 86\%   &99\% &91\%  & 99\% &90\% \\
\textbf{Ours}   & \textbf{97\%}   & \textbf{88\%}   &\textbf{99\%} &\textbf{94\%}  &\textbf{99\%} &\textbf{93\%}    \\ \bottomrule
\end{tabular}}
\label{table:singularity}
\vspace{-8mm}
\end{table}
\normalsize

% \paragraph{Experiment Setup} 
In this section, we first conduct experiments to evaluate the effectiveness of our design component in \textit{QuadraLib}, including the new quadratic neuron design and hybrid-BP based quadratic optimizer. Overall experiments are then conducted on several representative deep learning tasks (Image Classification, Object Detection, GAN-based Image Generation) to demonstrate the generality and scalability of \textit{QuadraLib} in various of tasks.
%
% \noindent \textbf{\textit{DNN Models and Datasets:}}
For Image Classification, three model structures are used as testing targets, including \textit{VGG-16}~\cite{simonyan2014very}, \textit{ResNet-32}~\cite{he2016deep}, 
% \textit{ResNet-50}, 
\textit{MobileNetV1}~\cite{howard2017mobilenets}.
Meanwhile, three datasets are included: \textit{CIFAR-10}, \textit{CIFAR-100}, and \textit{Tiny-ImageNet}.
% , and \textit{ILSVRC-2012};
For Image generation, one baseline network structure, namely, \textit{SNGAN}~\cite{miyato2018spectral} is evaluated on \textit{CIFAR-10}. We then convert original \textit{SNGAN} into our QDNN and compare its performance with baseline and other state-of-the-art works.  
For Object Detection, Single Shot MultiBox Detector (SSD)~\cite{liu2016ssd} framework with \textit{ PASCAL VOC2007} and \textit{PASCAL VOC2012} dataset are used as test model and corresponding dataset.

\subsection{Design Components Evaluation}

\noindent \textit{\textbf{Convergence Benefits of New Neuron Design:}}
We first evaluated our new neuron design. The results are shown in Table.~\ref{table:singularity}. 
Specifically, we use $\mathbb{T}2$, $\mathbb{T}3$, $\mathbb{T}4$ and $\mathbb{T}4$$+$Identity ($W_aXW_bX + X$) as baselines\footnote{$\mathbb{T}1$ is not practical to deployment with deep network structure due to high memory cost.} and tested different neuron designs in \textit{VGG-8}, \textit{VGG-16}, and \textit{ResNet-32} on \textit{CIFAR-10}.
Based on the results, we can found that for plain structures (VGGs), \textbf{\textit{$\mathbb{T}2$,$\mathbb{T}3$,$\mathbb{T}4$ fail to achieve convergence once the depth is increased from 8 to 16}}. 
Only by adding identity mapping or liner term in the quadratic neuron can help the model to achieve convergence. 

\begin{figure}[!t]
	\centering
	\captionsetup{justification=centering}
	\vspace{1mm}
	\includegraphics[width=3.3in]{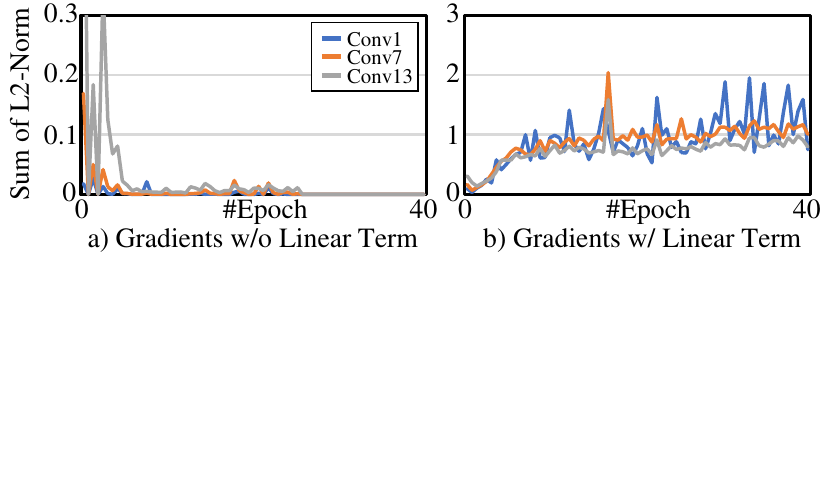}
	\vspace{-31mm}
	\caption{Gradient Evaluation for QDNN with \textit{VGG-16} Structure}
	\vspace{-1mm}
	\label{singularity}
\end{figure}

By contrast, our neuron design with linear term shows consistently better convergence performance with higher accuracy. 
Moreover, Fig.~\ref{singularity} shows the gradient value comparison for $\mathbb{T}3$ design without/with linear term, which is measured by our gradient distribution visualization toolset. We can easily find: without linear term or identity mapping function, gradients in the shallow layer (Conv1) quickly approach to zeros in the first few epochs and thereby weight parameters stop updating.
    By contrast, our neuron design by adding the linear term solves the gradient vanishing issue, enabling gradients to keep as meaningful values during training, thus achieving better training effectiveness.

\begin{figure}[t]
	\centering
	\captionsetup{justification=centering}
% 	\vspace{-2mm}
	\includegraphics[width=3.3in]{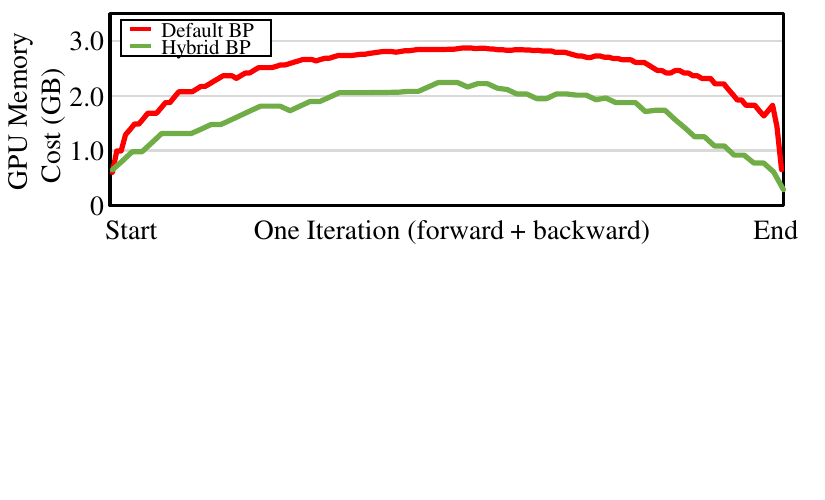}
	\vspace{-31mm}
	\caption{GPU Memory Cost Results with Hybrid-BP}
	\vspace{-4mm}
	\label{memory_optimization}
\end{figure}

\vspace{-1mm}
\noindent \textit{\textbf{Memory Savings of Hybrid-BP Quadratic Optimizer:}}
We evaluate the performance of our quadratic optimizer by hybrid back-propagation on a ConvNet  (3 conv layres and 2 full-connection layers). 
% The training dataset is \textit{CIFAR-10} and 
The batch size is 256 and the input size is 32$\times$32. We leverage $\textit{torch.cuda.memory\_allocated()}$ to measure the GPU memory occupied by QDNN training including forward and backward. The evaluation results are shown in Fig.~\ref{memory_optimization}. 
First, the memory cost during forward is constantly increased (indicated by red line) since intermediate parameters are gradually stored and they will be releases from GPU memory during backward. The required minimum GPU memory space is around 3GB. By contrast, when using our hybrid back-propagation (indicated by green line), the required GPU memory cost decreases to 2.2 GB, which shows a 26.7\% memory cost saving and higher training memory efficiency.

% First, we can find the memory cost is constantly increasing (indicated by red line) during forward since intermediate parameters are gradually stored from the first layer to the last layer. However, once the corresponding gradients are calculated, the stored intermediate parameters will be release from GPU memory during backward. Therefore, the required minimum GPU memory space is around 3GB. By contrast, when using our hybrid back-propagation (indicated by green line), the required GPU memory cost decreases to 2.2 GB, which shows a 26.7\% memory cost saving and higher training memory efficiency.  

\vspace{-0.5mm}
\subsection{Evaluation for Image Classification}
\vspace{-0.5mm}

\begin{table*}[t]
\small
\renewcommand\arraystretch{1.0}
\centering
\vspace{0mm}
\caption{The Performance Evaluation on \textit{CIFAR}}
\vspace{2mm}
\setlength{\tabcolsep}{0.9mm}{
\begin{tabular}{@{}cccccccccc@{}}
\toprule
\multicolumn{2}{c}{Model} & \#Layer/\#Block & \#Param & \begin{tabular}[c]{@{}c@{}}Train \\ Time/Batch \end{tabular} & \begin{tabular}[c]{@{}c@{}}Train \\ Memory\end{tabular} & \begin{tabular}[c]{@{}c@{}}Test \\ Time/Batch \end{tabular} & \begin{tabular}[c]{@{}c@{}}Accuracy\\ (CIFAR-10)\end{tabular} & \begin{tabular}[c]{@{}c@{}}Accuracy\\ (CIFAR-100)\end{tabular} \\ \midrule
\multirow{4}{*}{VGG-16}       & First-order$\dagger$                     &13 CL$^{\ast}$ &1.47E+7 &28ms &4.4GB &9ms   & 93.01\%  &73.11\%    \\
                              & ~\cite{fan2018new}           &7 CL  &1.20E+7  &56ms &3.3GB &19ms &92.48\% &72.47\% \\
                              & ~\cite{bu2021quadratic}      &7 CL    &0.80E+7 &43ms &3.2GB & 14ms &93.22\%  & 72.33\% \\
                              & QuadraNN (no auto-builder)     &13 CL  & 4.41E+7  &93ms &6.3GB &42ms   & 93.31\%  &73.23\%  \\
                              & QuadraNN                           &7 CL  & 1.20E+7  &53ms &3.3GB &18ms   & \textbf{94.03\%}  &\textbf{73.72\%}  \\ \cmidrule(l){1-9}
\multirow{4}{*}{ResNet-32}    & First-order                     &BS$^{\star}$:{[}5, 5, 5{]}  &4.80E+5  &28ms &2.9GB  &11ms  & 91.80\%  &68.59\%  \\
                              & ~\cite{fan2018new}         &BS:{[}2, 2, 2{]}  &3.93E+5 &33ms  &2.4GB &16ms &91.92\%  &67.83\%  \\
                              & ~\cite{bu2021quadratic}      &BS:{[}2, 2, 2{]}  &2.62E+5 &31ms &2.1GB &14ms &91.23\%  &67.70\%  \\
                              & QuadraNN (no auto-builder)     &BS: {[}5, 5, 5{]}  & 14.24E+7  &89ms &5.3GB &33ms   & 91.98\%  &69.98\%  \\
                              & QuadraNN                           &BS:{[}2, 2, 2{]}  &3.93E+5  &32ms &2.2GB  &15ms  &\textbf{92.38\%}  &\textbf{70.10\%}  \\ \cmidrule(l){1-9}
\multirow{4}{*}{MobileNet-V1} & First-order                     & 13 DW$^{\ddagger}$  &4.22E+6 &27ms  &4.8GB &10ms  &\textbf{92.97\%}  &\textbf{70.35}\%  \\
                              & ~\cite{fan2018new}       &8 DW  &3.53E+6  &35ms  &3.6GB &11ms  &91.59\% &68.31\%  \\
                              & ~\cite{bu2021quadratic}         & 8 DW  &2.97E+6  &25ms &3.2GB  &8ms  &92.18\% &69.77\% \\
                              & QuadraNN (no auto-builder)     &13 DW  & 12.66E+7  &112ms &7.6GB &47ms   & 91.81\%  &69.93\%  \\
                              & QuadraNN                        &8 DW &3.53E+6  &32ms  &3.3GB  &10ms  &92.44\%  & 70.03\% \\ \bottomrule
\end{tabular}}
\scriptsize
\leftline{\hspace{2mm}First-order$^{\dagger}$: the original \textit{VGG-16} with linear neurons. CL$^{\ast}$: Convolution layers. DW$^{\ddagger}$: a pair of depth-wise and point-wise convolution layers.}
\leftline{\hspace{2mm} BS$^{^{\star}}$: block numbers. For example,  {[}5, 5, 5{]} means the network has 3 blocks and each block includes 5 residual connections.}
% \leftline{\hspace{3mm} Design 1 and Design 2 indicate~\cite{fan2018new} and~\cite{bu2021quadratic}, respectively.}
\label{table:cifar}
\vspace{-5mm}
\end{table*}
\normalsize

\noindent \textit{\textbf{Experiment Setup:}}
For all three datasets, we use SGD optimizer and CosineAnnealing scheduler~\cite{loshchilov2016sgdr} with an initial learning rate of 0.1 and the training epoch is set as 200. The batch size for \textit{CIFAR} and \textit{Tiny-ImageNet} is set as 256 and 128, respectively. 
% For \textit{ILSVRC-2012}, we use momentum SGD and the learning rate decreases 10 times every 30 epochs with an initial learning rate of 0.1. The batch size is set as 16 for both training and 32 for inference. 
The first-order DNNs are used as baselines and two other state-of-the-art QDNN designs are evaluated, which are indicated as Design 1~\cite{fan2018new} and Design 2~\cite{bu2021quadratic}. The model generated from the proposed auto-builder is name as ``\textbf{QuadraNN}''. In order to validate auto-builder's effectiveness, we naively replace the first-order DNN with quadratic neuron in each layer and name as ``QuadraNN (no auto-builder)''.  All tests are running 10 times and we calculate the mean values.

\begin{table}[t]
\small
\renewcommand\arraystretch{1.3}
\centering
\vspace{-2mm}
\caption{The Performance of \textit{VGGNet} on \textit{Tiny-ImageNet}}
\vspace{2mm}
\setlength{\tabcolsep}{3mm}{
\begin{tabular}{@{}ccc@{}}
\toprule
Model & \#Layer & Accuracy \\ \midrule
First-order  & 13 CL     & 62.38\%        \\
QuadraNN   & 7 CL    & \textbf{62.66\%}      \\
QuadraNN (no ReLu)  & 7 CL  & 59.43\%  \\ \bottomrule
\end{tabular}}
\vspace{1mm}
\scriptsize
% \leftline{\hspace{-1mm} IS: Inception Score. FID: Frechet Inception Distance.}
\label{table:tiny_imagenet}
\vspace{-5mm}
\end{table}
\normalsize

\begin{table}[t]
\small
\renewcommand\arraystretch{1.3}
\centering
\vspace{-2mm}
\caption{The QDNN Performance Evaluation for Image Generation}
\vspace{2mm}
\setlength{\tabcolsep}{3mm}{
\begin{tabular}{@{}ccc@{}}
\toprule
Model & IS($\uparrow$) & FID($\downarrow$) \\ \midrule
CQFG\cite{lucas2019adversarial}  & 8.10                        & 18.60                          \\
EBM\cite{du2019implicit}   & 6.78                        & 38.2                           \\
SNGAN\cite{miyato2018spectral} & 8.06$\pm$0.10               & 19.06$\pm$0.50                 \\
PolyNet~\cite{chrysos2020p}  & 8.49$\pm$0.11               & 16.79$\pm$0.81                \\
QuadraNN  & \textbf{8.58$\pm$0.07}               & \textbf{16.53$\pm$0.73}                 \\ \bottomrule
\end{tabular}}
\vspace{1mm}
\scriptsize
\leftline{\hspace{-1mm} IS: Inception Score. FID: Frechet Inception Distance.}
\label{table:image_generation}
\vspace{-8mm}
\end{table}
\normalsize

\begin{figure}[t]
	\centering
	\captionsetup{justification=centering}
	\vspace{0.5mm}
	\includegraphics[width=3.3in]{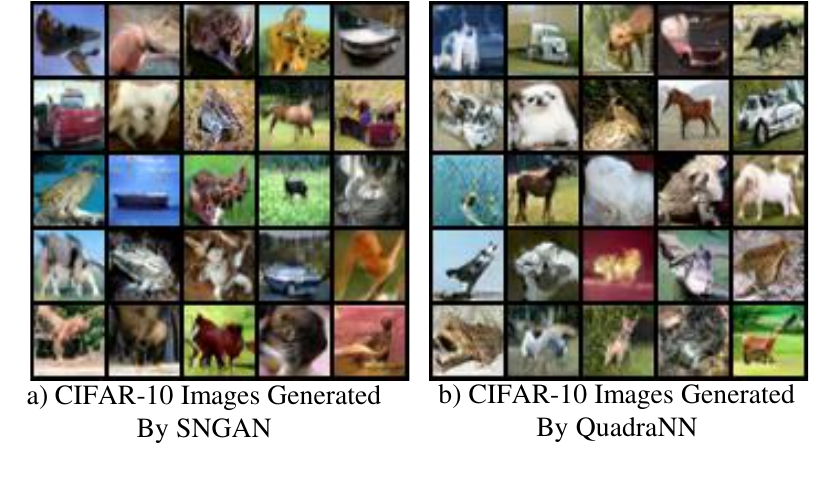}
	\vspace{-14.5mm}
	\caption{Image Generation Example}
	\vspace{-7mm}
	\label{image_generation}
\end{figure}

% \subsection {\textbf{Experiment Settings}}
% \vspace{-0mm}

% In this section, we first evaluate the end-to-end performance of \textit{QuadraLib} on three popular learning tasks: Image Classification, Image Generation, and Object Detection. We then validate the performance of each proposed design components with numerical experiments, including our unified neuron and quadratic optimizer by hybrid back-propagation. 

\begin{table*}[t]

\renewcommand\arraystretch{1.3}
\centering
\vspace{0mm}
\caption{The Detection Performance Evaluation on PASCAL VOC2007}
\vspace{2mm}
\small
\setlength{\tabcolsep}{1.6mm}{
% \begin{tabular}{c|c|c|c|c|c|c|c|c|c|c|c|c|c|c|c|c}
\begin{tabular}{ccccccccccccccccc}
\toprule
Model &Pre-trained&
  \begin{tabular}[c]{@{}c@{}}plane\end{tabular} &
  bike &
  bird &
  boat &
  bottle &
  bus &
  car &
  cat &
  chair &
  $\cdots$ &
  sheep &
  sofa &
  train &
  TV &
  \textbf{Total} \\ \hline
1st order &$\times$
   &0.63
   &0.61
   &0.33
   &0.37
   &0.12
   &0.64
   &0.73
   &0.59
   &0.27
   &$\cdots$
   &0.51
   &0.52
   &0.67
   &0.53
   &0.52
   \\ \hline
QuadraNN &$\times$
   & 0.76
   & 0.80
   & 0.66
   & 0.65
   & 0.36
   & 0.82
   & 0.84
   & 0.84
   & 0.53
   & $\cdots$
   & 0.70
   & 0.77
   & 0.86
   & 0.71
   & \textbf{0.73}
   \\ \hline
   1st order &$\surd$
   & 0.78
   & 0.83
   & 0.76
   & 0.72
   & 0.45
   & 0.87
   & 0.86
   & 0.88
   & 0.59
   & $\cdots$
   & 0.79
   & 0.79
   & 0.86
   & 0.74
   & 0.76
   \\ \hline
QuadraNN &$\surd$
   & 0.78
   & 0.84
   & 0.78
   & 0.73
   & 0.47
   & 0.88
   & 0.86
   & 0.89
   & 0.63
   & $\cdots$
   & 0.79
   & 0.82
   & 0.89
   & 0.76
   & \textbf{0.78}
   \\ \bottomrule
\end{tabular}}
\scriptsize
% \leftline{\hspace{-1mm} IS: Inception Score. FID: Frechet Inception Distance.}
\label{table:object_detection}
\vspace{-6mm}
\end{table*}
\normalsize

\noindent \textit{\textbf{Results Analysis:}}
The evaluation results are shown in Table.~\ref{table:cifar}. We can found that our design (QuadraNN) can always show better accuracy performance than baseline and other designs on both \textit{CIFAR-10} and \textit{CIFAR-100}. Specifically, compared to the original first-order DNN with \textit{VGG-16} structure, QuadraNN only needs 7 convolution layers, which achieves 18.36\% parameter size and 11.30\% GPU memory cost reduction. Meanwhile, QuadraNN shows 1.01\% and 0.61\% accuracy improvement. Moreover, if we directly generate QDNN without applying converter, the model size is significant large with higher consumption and latency. When using converter, we can achieve 4$\times$, 3$\times$, 1.5$\times$ saving in terms of parameter size, training time cost, and memory occupancy.
For \textit{ResNet-32} structure, QuadraNN only need 2 skipping connections structure in each residual block, thereby decrease 23.90\% parameter size and 6.89\% GPU memory cost. On the contrary, QuadraNN shows 0.58\% and 1.51\% accuracy improvement. Similarly, using our converter can achieve 4$\times$, 5$\times$, and 1.6$\times$ saving in terms of parameter size, training time cost, and memory occupancy.
For \textit{MobileNet-V1} structure, QuadraNN only needs 8 pair of depth-wise/point-wise convolution layers (DWs) while the original first-order DNN needs 13 DWs. Therefore, QuadraNN achieves 31.25\% GPU memory saving while achieve a comparable accuracy performance than the first-order DNN. Compared to QuadraNN (no converter), our model also shows better performance with 4$\times$, 3$\times$, and 2$\times$ saving in terms of parameter size, training time cost, and memory occupancy. 
Moreover, when increasing the image size, QuadraNN still can achieve the better performance trade-off between efficiency and accuracy than the first-order DNNs, which is shown in Table.~\ref{table:tiny_imagenet}.

\begin{figure}[t]
	\centering
	\captionsetup{justification=centering}
	\vspace{1mm}
	\includegraphics[width=3.3in]{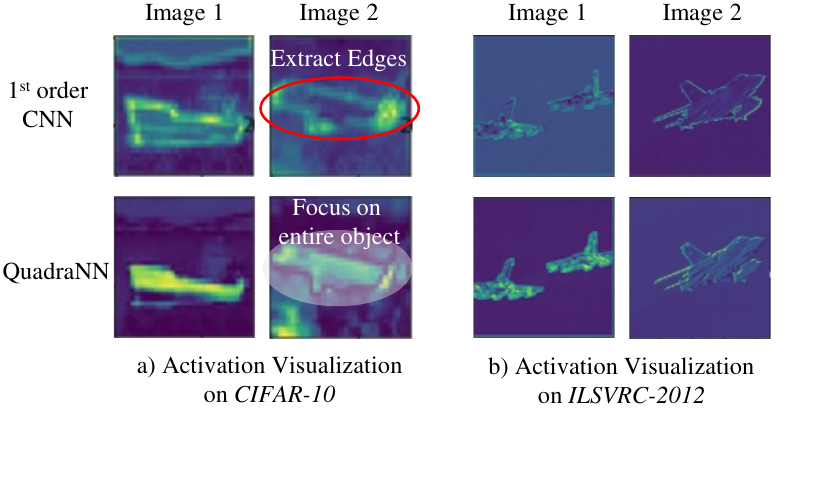}
	\vspace{-17mm}
	\caption{Activation Attention via Our Visualization Tool}
	\vspace{-8mm}
	\label{activation_visulization}
\end{figure}

\vspace{0.7mm}
\subsection{Evaluation for Image Generation}

\vspace{0.5mm}
\noindent \textit{\textbf{Experiment Setup:}}
We adopt \textit{SNGAN}~\cite{miyato2018spectral} as the baseline structure and test it on \textit{CIFAR-10}. The original generator in \textit{SNGAN} has 3 residual blocks. For our QuadraNN, we convert each convolution layer in generator into our quadratic layer. The structure of discriminator and other hyper-parameters are same as settings in~\cite{miyato2018spectral}. Three others state-of-the-art works are compared~\cite{lucas2019adversarial, du2019implicit, chrysos2020p}. Similar to~\cite{chrysos2020p}, each network was run for 10 times and their mean and variance are recorded. We use Inception Score (IS)~\cite{salimans2016improved} and Frechet Inception Distance (FID)~\cite{heusel2017gans} as performance metrics. A higher IS score and a lower FID score indicate a better generation performance. 

\vspace{-0.5mm}
\noindent \textit{\textbf{Results Analysis:}}
The results are shown in Table.~\ref{table:image_generation}. Compared to the baseline \textit{SNGAN} and other first-order DNNs, QDNN models (PolyNet and Ours) demonstrate a higher IS but lower FID scores, which indicates that they have performance superiority. Moreover, since our quadratic neuron design has a theoretical higher approximation capability, thereby shows a better performance than PolyNet. 
% Fig.~\ref{image_generation} shows the generated images from the original \textit{SNGAN} and our QuadraNN. We can found that the images generated by QuadraNN have more details and are more realistic than the images generated from \textit{SNGAN}.

\subsection{Evaluation on Object Detection}

\noindent \textit{\textbf{Experiment Setup:}} As aforementioned, we use SSD as object detector. The backbone of SSD is \textit{VGG-16}. 
    We use VOC2007 trainval and VOC2012 trainval as training set and use VOC2007 as test set. The initial learning rate is $10^{-3}$ and will decrease 10 times at iteration {[}80000, 100000 {]}.
In the original SSD, backbone network \textit{VGG-16} use a pre-trained model parameters from \textit{ILSVRC-2012}. In order to comprehensively evaluate the detection performance of QuadraNN, we have two settings: 1) We initialize both first-order and second-order \textit{VGG-16} with Kaiming initialization~\cite{he2015delving} without pre-training. 
2) We initialize the model parameters by copying the parameter from pre-trained QuadraNN on \textit{ILSVRC-2012}.

\noindent \textit{\textbf{Results Analysis:}}
Table.~\ref{table:object_detection} shows the evaluation results. For situation of training without pre-training, QuadraNN demonstrate approximate 50\% detection performance improvement than the ordinary first-order DNN. Such performance can even compete the first-order DNN with pre-training model parameters. For pre-training setting, QuadraNN still shows a better detection performance with 0.02 mAP improvement. 
% We believe this is because quadratic neuron can focus on the objects themselves instead of edges, which we discussed in Section V.  Therefore it has better feature extraction ability and is especially critical for object detection task. 
We believe the superiority of QDNN in object detection is due to the difference characteristic of feature extraction between first-order layer and second-order layer.  
In order to validate our assumption, we leverage our activation attention visualization tool to compare the feature extraction results for the first layer on \textit{CIFAR-10} and \textit{ILVSRC-2012}, which is shown Fig.~\ref{activation_visulization}. 
We can get an interesting finding: the traditional first-order convolutional layer more focuses on edges, including edges of objects and background. 
On the contrary, quadratic layer can \textit{\textbf{extract the entire objects}} (shown as white circle area). This feature extraction characteristic will highly contribute to object detection task.

	\vspace{-1.5mm}
\section{Conclusion and Discussion}
\label{sec:conc}
	\vspace{-0.8mm}

In this paper, we comprehensively reviewed the existing QDNN architecture and proposed a new quadratic neuron design with theoretical effectiveness and efficiency analysis. We further developed ``\textit{QuadraLib}'', a QDNN library to support QDNN design with model structure construction and training optimization. The extensive experiments from multiple learning tasks demonstrate that our designed QDNN has superiority compared to the first-order DNNs and other existing QDNN works.
Furthermore, from our theoretical analysis and numerical experiments, we believed that QDNNs show a significant potential on learning tasks which highly depends on extracted objects, such as object detection, segmentation, and position recognition. 
By applying QDNNs on these scenarios, the recognition process will more focus on the important objects while ignoring the unimportant background, thus is expected to achieve a better performance improvement gain. 
% Therefore, in our future work, we will explore more comprehensive QDNN design on this kind of learning tasks.

% In this work, we proposed \textit{Helios} --- a heterogeneous-aware FL framework with dynamically balanced collaboration.
% Leveraging on-device model training consumption profiling and the innovative ``soft-training'' training scheme, \textit{Helios} can introduce effective local CNN model optimization into FL to eliminate stragglers, while maintaining expected collaborative convergence across all edge devices.
% Experiments demonstrated that the proposed \textit{Helios} achieves superior training accuracy, speed, as well as Non-IID setting resistance.
% By well addressing the computational heterogeneity of edge devices, the proposed \textit{Helios} significantly enhanced the applicability and performance of federated learning for training-on-edge.

% \begin{figure}[!t]
% 	\centering
% 	\captionsetup{justification=centering}
% 	\vspace{-7mm}
% 	\includegraphics[width=3.3in]{noniidexp}
% 	\vspace{-11.5mm}
% 	\caption{\textit{Helios} Evaluation with Non-IID Data}
% 	\vspace{-5.5mm}
% 	\label{uneqnon}
% \end{figure}

\newpage
\bibliography{Zirui_MLSys2022_QuadraLib}
\bibliographystyle{mlsys2022}

\end{document}